\documentclass[letterpaper]{article} % DO NOT CHANGE THIS
\usepackage{aaai2026}  % DO NOT CHANGE THIS
\usepackage{times}  % DO NOT CHANGE THIS
\usepackage{helvet}  % DO NOT CHANGE THIS
\usepackage{courier}  % DO NOT CHANGE THIS
\usepackage[hyphens]{url}  % DO NOT CHANGE THIS
\usepackage{graphicx} % DO NOT CHANGE THIS
\usepackage{natbib}  % DO NOT CHANGE THIS AND DO NOT ADD ANY OPTIONS TO IT
\usepackage{caption} % DO NOT CHANGE THIS AND DO NOT ADD ANY OPTIONS TO IT
\usepackage{algorithm}
\usepackage{algorithmic}

\usepackage{newfloat}
\usepackage{listings}
\DeclareCaptionStyle{ruled}{labelfont=normalfont,labelsep=colon,strut=off} % DO NOT CHANGE THIS
\floatstyle{ruled}
\newfloat{listing}{tb}{lst}{}
\floatname{listing}{Listing}
\usepackage{booktabs}
\usepackage{amsmath, amsfonts}
\usepackage[most]{tcolorbox}

\title{DeepWriter: A Fact-Grounded Multimodal Writing Assistant Based On Offline Knowledge Base}
\author{
    Song Mao\textsuperscript{\rm 1}\equalcontrib,
    Lejun Cheng\textsuperscript{\rm 2}\equalcontrib\thanks{work done during the internship at Shanghai AI LAB.},
    Pinlong Cai\textsuperscript{\rm 1}\thanks{Corresponding Author.},
    Guohang Yan\textsuperscript{\rm 1},
    Ding Wang\textsuperscript{\rm 1},
    Botian Shi\textsuperscript{\rm 1}
}
\affiliations{
    \textsuperscript{\rm 1}Shanghai Artificial Intelligence Laboratory\\
    \{maosong, caipinlong, yanguohang, wangding, shibotian\}@pjlab.org.cn

    \textsuperscript{\rm 2}Peiking University\\
    chenglj023@gmail.com
}

\usepackage{bibentry}
\begin{document}

\maketitle

\begin{abstract}
Large Language Models (LLMs) have demonstrated remarkable capabilities in various applications. However, their use as writing assistants in specialized domains like finance, medicine, and law is often hampered by a lack of deep domain-specific knowledge and a tendency to hallucinate. 
Existing solutions, such as Retrieval-Augmented Generation (RAG), can suffer from inconsistency across multiple retrieval steps, while online search-based methods often degrade quality due to unreliable web content. 
To address these challenges, we introduce \textbf{DeepWriter}, a multimodal, long-form and fact-grounded writing assistant that operates on a curated, offline knowledge base.
DeepWriter leverages a novel pipeline that involves task decomposition, outline generation, multimodal retrieval, and section-by-section composition with reflection. 
By deeply mining information from a structured corpus and incorporating both textual and visual elements, DeepWriter generates coherent, factually grounded, and professional-grade documents. 
To evaluate the performance of DeepWriter, we curate a benchmark containing five domains, experiment results on the curated benchmark demonstrate that DeepWriter produces high-quality, verifiable articles that surpasses existing baselines in factual accuracy and generated content quality.
\end{abstract}


\section{Introduction}
The rapid advancement of Large Language Models (LLMs) and Multimodal Large Language Models (MLLMs) has unlocked a wide array of applications, from programming and education to complex task automation. 
Commercial models like Gemini~\cite{geminipro2.5}, Claude 3.5~\cite{claude3series2024} and GPT-4o~\cite{chatgpt4o} have shown proficiency in leveraging web search and deep reasoning to assist users with real-world problems.

In this paper, we investigate the application of LLMs as sophisticated writing assistants. While the context windows of modern LLMs are sufficient for processing entire novels~\cite{team2024gemini, minimax2025minimax01scalingfoundationmodels, yang2025qwen251mtechnicalreport}, their application in professional writing often remains at a superficial level of data processing and summarization~\cite{wu2025shiftinglongcontextllmsresearch}. 
In specialized domains such as finance, medicine, and law, LLMs frequently fail to generate expert-level responses, and are prone to factual hallucinations. 
Two primary paradigms have emerged to mitigate these issues. 
The first, Retrieval-Augmented Generation (RAG)~\cite{lewis2020retrieval}, connects LLMs to external knowledge bases.
However, RAG can struggle with maintaining coherence and consistency across long-form documents, as multi-turn retrieval may introduce disjointed information~\cite{gu2025rapidefficientretrievalaugmentedlong}. 
The second paradigm involves using online search to gather relevant information~\cite{openai2025deepresearch, geminideepresearch, grokdeepresearch, comet-perplexity-2025}, but the variable quality of web content can compromise the overall quality and reliability of the generated document.

\begin{figure}[!tb]
  \centering
  \includegraphics[width=\linewidth]{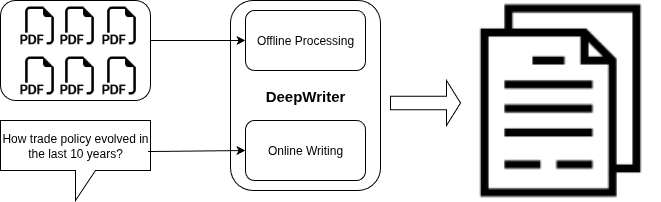}
  \caption{\textbf{Illustration of DeepWriter}. DeepWriter receives a query and a offline corpus to produce a long-form, multimodal and fact-grounded article.}
  \label{fig:deepwriter_illustration}
\end{figure}

To overcome these limitations, we propose \textbf{DeepWriter}, a multimodal, long-context writing assistant that operates on a curated, offline corpus. 
Unlike systems that primarily gather information from structured web pages, DeepWriter is engineered to deeply mine knowledge from information-rich but unstructured sources like PDFs. 
Our pipeline utilizes advanced document processing toolkits to extract and meticulously organize text, tables, and images with their corresponding metadata.

To integrate retrieved materials structurally,  we propose a multi-stage pipeline, consisting of task decomposition, document clustering, section by section writing. 
Previous work have shown that such a multi-stage procedure can achieve superiror performance compared with single stage pipeline~\cite{bai2025longwriter, wu2025superwriterreflectiondrivenlongformgeneration}. 
By utilizing a multi-stage processing pipeline, we mitigate the overload of integrating external knowledge and internal knowledge. 

Next, we believe that multimodal contents is more appealing to users compared to pure text reports, thus we propose a novel multimodal reranking system to intelligently determine the most appropriate location for images and charts within the text. 
This foundation allows DeepWriter to structurally organize the retrieved information for multimodal generation. 

Finally, to ensure the final output is fully verifiable and trustworthy, DeepWriter excels at generating and citing multimodal content with fine-grained granularity. 
Instead of citing an entire source file, which is a common practice that makes validation intractable for large documents, our system tracks and attributes information at the paragraph level. 
This approach guarantees that the generated document is not only coherent and professional but also factually grounded and reliable, directly addressing the core challenge of hallucination in LLMs.

We test DeepWriter on report generation task corresponding to five domains.
Experiment results show that DeepWriter, with a simple yet effective framework, can achieve comparable performance with other leading open-source long-context models.

The key contributions of our work include:

\begin{enumerate}
    \item \textbf{Hierarchical Knowledge Representation}: We proposed a three-level (knowledge-chunk-document) representation that enables efficient multi-granularity retrieval while maintaining precise source attribution capabilities.
    
    \item \textbf{Structured Writing Pipeline}: We developed a comprehensive pipeline that includes query rewriting, task decomposition, outline generation, iterative retrieval, section-by-section writing, and reflection mechanisms to ensure quality and coherence.
    
    \item \textbf{Multimodal Content Integration}: We introduced an interleaved image-text generation approach with relevance scoring and placement optimization algorithms that seamlessly integrate visual and textual content.
    
    \item \textbf{Fine-grained Citation System}: We implemented a three-level citation system (document, paragraph, sentence) that provides precise source attribution and enables comprehensive fact verification.
\end{enumerate}

\section{Related Work}
Our work is situated at the intersection of three research areas: AI-powered research assistants, automated writing assistants, and Retrieval-Augmented Generation (RAG).

\paragraph{Research Assistants}
Several projects have explored the integration of LLMs with public data sources for scientific question-answering and writing. 
Systems like OpenScholar~\cite{asai2024openscholarsynthesizingscientificliterature} and ScholarQA~\cite{singh2025ai2scholarqaorganized} aim to assist researchers by surveying and synthesizing information.
DeepResearcher~\cite{zheng2025deepresearcherscalingdeepresearch} trained a LLM-based deep research agents end-to-end through scaling reinforcement learning in real-word environments.
DeepWriter extends this line of work by focusing on long-form, professional document generation from a controlled, offline corpus, with an emphasis on multimodal content generation.

\paragraph{Retrieval-Augmented Generation (RAG)}
RAG is a powerful technique for grounding LLM outputs in external knowledge, addressing issues of outdated information and hallucination. Innovations like GraphRAG~\cite{edge2025localglobalgraphrag}, LightRAG~\cite{guo2024lightrag}, and KAG~\cite{liang2024kag} have introduced more sophisticated retrieval mechanisms. 
However, generating long, coherent documents with RAG remains a challenge. 
DeepWriter's structured pipeline and hierarchical knowledge representation are designed to improve coherence and factual consistency in long-form generation tasks.

\paragraph{Writing Assistants}
The concept of an AI writing assistant is popular, with a focus on improving human productivity by generating text in various styles.
Frameworks like OmniThink~\cite{xi2025omnithinkexpandingknowledgeboundaries} have explored a ``slow-thinking" approach to generate articles based on search engine results. However, OmniThink did not address multimodal information and relies on web search capability. 
Other related systems like STORM~\cite{shao-etal-2024-assisting} and CO-STORM~\cite{jiang2024unknownunknownsengagedhuman} focus on generating outlines and then ``filling in the blanks" but rely on web searches, which can introduce noise, besides, the generated article are constrained in Wikipedia format. 
RAPID can generate comprehensive and knowledge-intensive articles based on RAG~\cite{gu2025rapidefficientretrievalaugmentedlong}.
QRAFT is an LLM-based agentic framework that mimics the writing workflow of human fact-checkers~\cite{sahnan2025can}.
~\cite{xiong2025outliningheterogeneousrecursiveplanning} proposed a general agent framework that achieves human like adaptive writing through recursive task decomposition and dynamic integration of three fundamental task types, i.e. retrieval, reasoning, and composition.
SuperWriter~\cite{wu2025superwriterreflectiondrivenlongformgeneration} proposed a reflection-driven agent framework that decomposes long-form text generation into a planning, writing, and refining loop, enabling iterative self-improvement through hierarchical direct preference optimization.
DeepWriter differentiates itself by using a curated knowledge base and integrating multimodal content generation.

\section{Methodology}
\subsection{Task Definition}
Our goal is to generate a long document $P$ given a collected domain-specific corpus $\mathcal{K}=\{D_1, \dots, D_m\}$ and a user-provided query or topic $Q$,  with the following characteristics:

\begin{enumerate}
    \item \textbf{Multimodal Content}: The document should include relevant images, tables, and charts derived from the corpus to enrich the modality of the document.
    \item \textbf{Factual Grounding}: All claims, figures, and critical viewpoints must be accompanied by citations pointing to their source in the corpus to minimizing the hallucination bought by the LLMs.
    \item \textbf{Semantic Coherence}: The document must be logically consistent and flow naturally from one section to the next.
\end{enumerate}
Compared with existing writing assistants, DeepWriter aims to thoroughly put information together. 
Since prompting for innovation may cause hallucination, in this paper, we require the agent to give less or simply not pose any novel conclusion.

\subsection{System Architecture}
\begin{figure*}[t]
  \includegraphics[width=\textwidth]{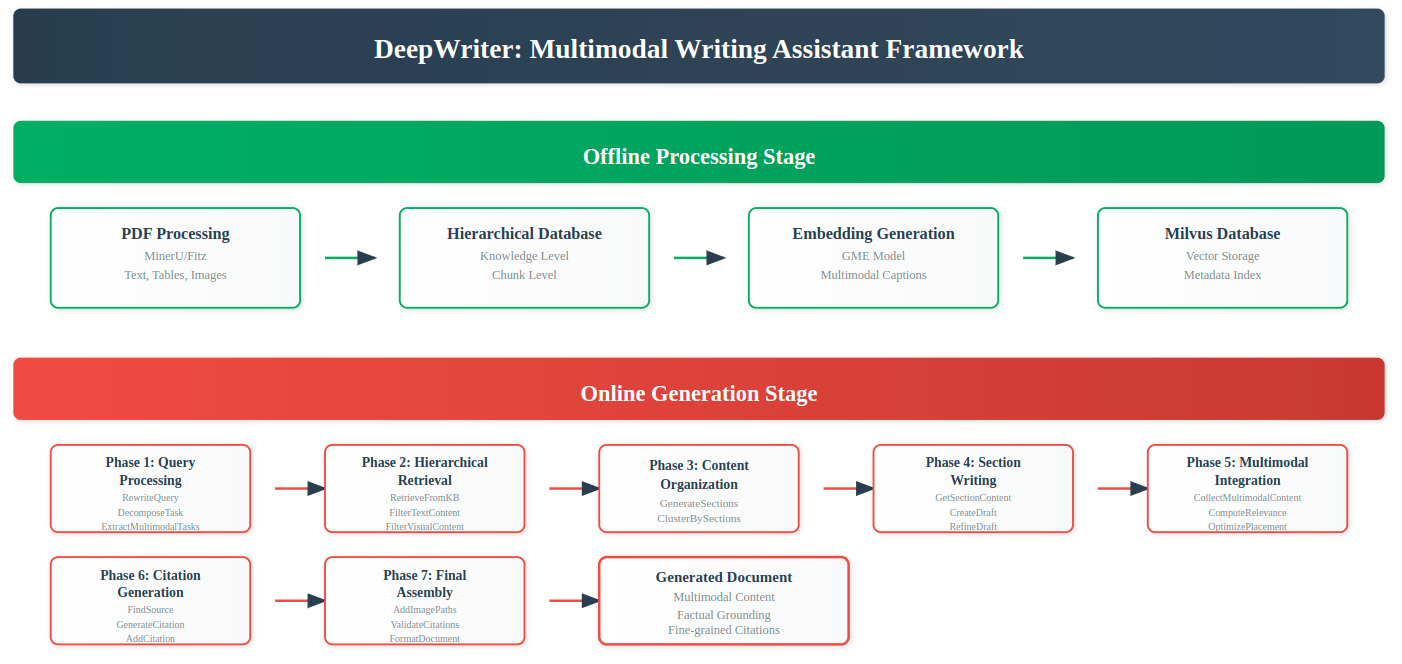}
  \caption{The overall framework of DeepWriter.}
  \label{fig:deepwriter_architecture}
\end{figure*}

The DeepWriter pipeline consists of two stages: an offline processing stage that is responsible for converting the unstructured PDFs into unified format for retrieval,  and an online writing stage, responsible for generating the final output, the framework of DeepWriter is as illustrated in the Fig~\ref{fig:deepwriter_architecture}.

\paragraph{Offline Processing Stage}
During the offline stage, we aim to preprocess corpus and store them in an easy-to-retrieve format in a knowledge base $\mathcal{B}$.

In the data processing stage, we aim to extract structured information from unstructured file formats such as PDF. 
We first use Fitz\footnote{\url{https://github.com/pymupdf/PyMuPDF}} or MinerU~\cite{wang2024mineruopensourcesolutionprecise} to extract text, tables, images from the original unstructured documents. 
During the process, we keep all file and page metadata to avoid information loss when performing fact-checking.
After extracting information from files, we choose appropriate chunk size to avoid chunks that are too small or too large.
We also use advanced VLM such as Qwen2.5-VL~\cite{bai2025qwen2_5} to generate detailed image or table captions for multimodal retrieval.

\begin{figure}[!htb]
  \includegraphics[width=\columnwidth]{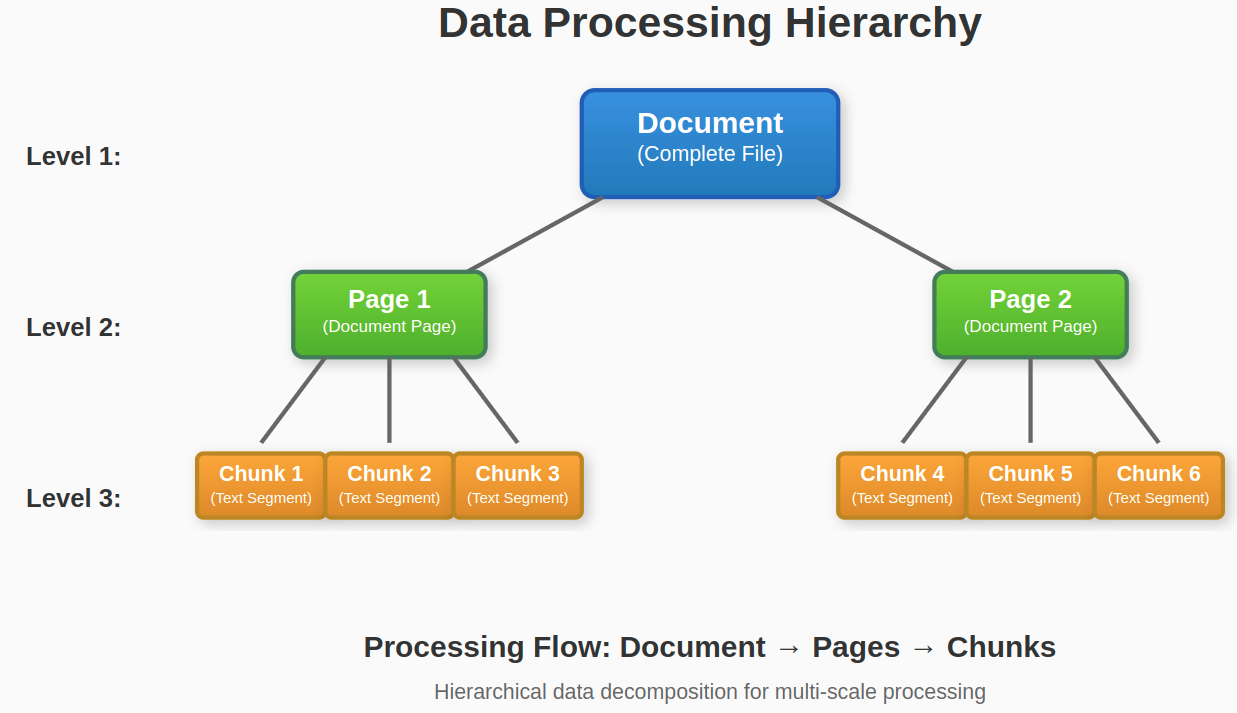}
  \caption{Hierarchical structure of data processing. The document is decomposed into pages, and each page is further divided into text chunks. This multi-level organization enables fine-grained and scalable data analysis.}
  \label{fig:data_processing_hierarchy}
\end{figure}

We structure the extracted information into a hierarchical database. This involves a three-level representation (chunk-page-document) to facilitate efficient, multi-granularity retrieval.
The top ``document" level contains abstract concepts, such as year or domain information, the ``page" level is an intermediate level, which trades offs the efficiency and granularity of retrieval, the final ``chunk" level is composed of knowledge pieces, holding the raw source information. The data processing hierarchy is shown in Fig~\ref{fig:data_processing_hierarchy}.

\begin{algorithm}[!htb]
\caption{DeepWriter Online Generation Stage Pipeline}
\label{alg:online_generation}
\begin{algorithmic}[1]
\REQUIRE User query $Q$, Offline knowledge base $\mathcal{K}$
\ENSURE Generated document $P$ with multimodal content and citations

\STATE \textbf{Phase 1: Query Processing}
\STATE $Q' \leftarrow \text{RewriteQuery}(Q)$ 
\STATE $\mathcal{T} \leftarrow \text{DecomposeTask}(Q')$ 

\STATE \textbf{Phase 2: Hierarchical Retrieval}
\FOR{each subtask $t_i \in \mathcal{T}$}
    \STATE $\mathcal{R}_i \leftarrow \text{RetrieveFromKB}(t_i, \mathcal{K})$
    \STATE $\mathcal{R}_i^{\text{text}} \leftarrow \text{FilterTextContent}(\mathcal{R}_i)$
    \STATE $\mathcal{R}_i^{\text{visual}} \leftarrow \text{FilterVisualContent}(\mathcal{R}_i)$
\ENDFOR

\STATE \textbf{Phase 3: Content Organization}
\STATE $\mathcal{S} \leftarrow \text{GenerateSectionTitles}(\mathcal{T})$ 
\STATE $\mathcal{C} \leftarrow \text{ClusterBySections}(\mathcal{R}^{\text{text}}, \mathcal{R}^{\text{visual}}, \mathcal{S})$ 

\STATE \textbf{Phase 4: Section-by-Section Writing}
\STATE $P \leftarrow \emptyset$, $\mathcal{H} \leftarrow \emptyset$

\FOR{each section $s_j \in \mathcal{S}$}
    \STATE $\mathcal{C}_j \leftarrow \text{GetSectionContent}(\mathcal{C}, s_j)$
    \STATE $d_j \leftarrow \text{CreateDraft}(s_j, \mathcal{C}_j, \mathcal{H})$
    \STATE $d_j' \leftarrow \text{RefineDraft}(d_j, \mathcal{C}_j)$
    \STATE $P \leftarrow P \cup \{d_j'\}$ 
    \STATE $\mathcal{H} \leftarrow \mathcal{H} \cup \text{SummarizeSection}(d_j')$
\ENDFOR

\STATE \textbf{Phase 5: Multimodal Integration}
\STATE $\mathcal{M} \leftarrow \text{CollectMultimodalContent}(\mathcal{R}^{\text{visual}})$
\FOR{each visual element $m_k \in \mathcal{M}$}
    \STATE $\text{score}_{k,\cdot} \leftarrow \text{ComputeRelevance}(m_k, P)$ 
    \STATE $\text{pos}_k \leftarrow \text{OptimizePlacement}(\text{score}_{k,\cdot}, P)$ 
    \STATE $P \leftarrow \text{InsertVisual}(P, m_k, \text{pos}_k)$ 
\ENDFOR

\STATE \textbf{Phase 6: Citation Generation}
\FOR{each claim $c_l$ in $P$}
    \STATE $\text{source}_l \leftarrow \text{FindSource}(c_l, \mathcal{R})$ 
    \STATE $\text{citation}_l \leftarrow \text{GenerateCitation}(\text{source}_l)$ 
    \STATE $P \leftarrow \text{AddCitation}(P, c_l, \text{citation}_l)$
\ENDFOR

\STATE \textbf{Phase 7: Final Assembly}
\STATE $P \leftarrow \text{AddImagePaths}(P)$ 
\STATE $P \leftarrow \text{ValidateCitations}(P)$ 
\STATE $P \leftarrow \text{FormatDocument}(P)$

\RETURN $P$
\end{algorithmic}
\end{algorithm}

\paragraph{Online Generation Stage}
During the online stage, we aim to generate a long-form, multimodal, factual grounded article based on user query $Q$ and the knowledge base $\mathcal{B}$ generated in offline stage.
The overall algorithm is introduced in Algorithm~\ref{alg:online_generation}.
We detail our key design in the following subsections.

\subsection{Multi-Stage generation pipeline}
Firstly, the initial user query $Q$ is rewritten and expanded to better reflect the user's intention, which mitigates ambiguity such as using abbreviations or general questions.
After rewriting the user's query, we decompose the task according to Fact, Data and Point.
The \emph{fact} contains new information or restricted information. For example, when talking about "What is the total trade volume of the world in 2024?", we need to specify the countries/regions that account for, which requires the agent to generate a sub-task on which countries should be included.
The \emph{data} directly quantifies the performance, trend or severity, which connects the fact and provides a strong foundation for the point.
The task decomposition module carefully asks questions to retrieve exact numerical values to improve reliability.
Besides text data, there are data illustrated in tables, charts or figures.
So, we also require the task decomposition module to design strategies for searching multimodal contents that are related to the topic.
The \emph{Point} are conclusion drawn by editors or authors, these conclusions are considered as a viewpoint as a reference.

Next, we perform retrieval according to decomposed sub-tasks or sub-queries based on the knowledge base.
To enable unified multimodal retrieval, we utilize a unified multimodal embedding model that supports images and texts as the inputs, this unified embedding model alleviates the burden of loading multiple model and improves the searching efficiency.
We then use cosine similarity as the metric to select top-$k$ relevant documents for each decomposed query.
The relevant documents are then clustered based on pre-generated section titles, which reduces information seeking cost when generating section contents.

Now, we perform the writing process for each section. We ask the agent to write section by section according to relevant documents, section title and content that is already written, focusing on combining the facts, data and points and writing in a general-to-specific pattern, such that audiences can focus on different parts, achieving customization capability.
To improve fluency, we require the agent to make a draft instead of writing directly. 
After completing the wriitng process, we ask the agent to summarized the written content to compress the context and prevent from generating repeated content.

Finally, we put everything together, including contents of each section, the multimodel contents and corresponding references. 
We integrate multimodal contents by compute the relevance among different modalities.
Then, we add citations for each section, and these citations are fine-grained such that we can trace to their original paragraphs. 

\subsection{Interleaved Image-Text Generation}
One challenge in generating multimodal articles lies in how to find proper places to insert multimodal contents.
For text contents, we can ask the LLM to quote it based on memory mechanism or the instruction following capability.
However, for multimodal contents such as images, tables, and charts, determining their optimal placement within the text requires sophisticated reasoning about content relevance and contextual flow.

To address this challenge, we propose a two-stage approach: content relevance scoring and contextual placement optimization. 
First, we compute relevance scores between each multimodal element and every paragraph in the generated text using the same embedding model. 
This creates a relevance matrix that captures semantic similarity between visual and textual content.

Second, we employ a placement optimization algorithm that considers both relevance scores and document flow constraints. 
The algorithm ensures that images are placed near their most relevant textual descriptions while maintaining the logical structure of the document. 
For tables and charts, we implement additional constraints to ensure they appear before or after the paragraphs that reference their data.
The algorithm is shown in Algorithm~\ref{alg:placement_optimization}.

\subsection{Grounded Citation}
A critical aspect of DeepWriter is its ability to provide precise, traceable citations that enable readers to verify claims and access source materials. 
Unlike traditional citation systems that reference entire documents, DeepWriter implements fine-grained citation that can point to specific pages, or even the paragraphs.

Our citation system operates at three levels of granularity: document-level, paragraph-level, and sentence-level. 
Each citation includes metadata such as source file name, page number, and chunk bounding box, allowing for precise source verification. 
For multimodal content, citations also include bounding box coordinates for images and tables.

The citation generation process is integrated into the writing pipeline, automatically creating citations as content is generated. 
This ensures that every factual claim, numerical value, or visual element is properly attributed to its source in the corpus. 
The system also maintains a citation consistency check to ensure that similar claims across different sections reference the same sources when appropriate.
The algorithm is shown in Algorithm~\ref{alg:citation_generation}.

\section{Experiments}

\subsection{System Configuration}
We use MinerU~\cite{wang2024mineruopensourcesolutionprecise} as PDF processing toolkit to process a bundle of PDFs. 
MinerU accepts multiple format files as input and output the OCR results in a markdown format. 
We store the extracted images and texts in a Milvus database, besides the processed information, we also add the embedding generated via GME (\texttt{gme-Qwen2-VL-2B-Instruct}\footnote{See \url{https://huggingface.co/Alibaba-NLP/gme-Qwen2-VL-2B-Instruct}})~\cite{zhang2025gmeimprovinguniversalmultimodal} as the retrieval key. 
For image and text items, a descriptive caption is generated via an advanced vision language model Qwen2.5-VL 7B~\cite{bai2025qwen2_5}.
We use Qwen2-7B~\cite{yang2024qwen2technicalreport} as the LLM performing online generation tasks.

% TODO: configuration for online generation

\subsection{Evaluation Setting}
\paragraph{Models}
We compare DeepWriter against the following models:

\begin{enumerate}
    \item Long-Context MLLMs: We use state-of-the-art LLMs with long context windows to assess their zero-shot long-form writing capabilities. 
    We use Qwen-Plus as the LLM to be compared in this paper, of which context window is 131,072 with maximum output of 8,192 tokens\footnote{See the official website ~\url{https://www.alibabacloud.com/help/en/model-studio/what-is-qwen-llm} for more details.}.
    \item Naive RAG: A standard RAG implementation to evaluate its performance on long-document generation. In this setting, we directly retrieve the relevant information based on original query, then ask the LLM to write an article based on retrieved information.
    \item Specialized Writing Models: We also compare the performance of DeepWriter against other long-form article generation systems, including STORM, CO-STORM.
\end{enumerate}
The comparisons for different systems is shown in Table~\ref{tab:system_comparison}.

\begin{table*}[h]
    \centering
    \caption{Configuration for different systems}
    \label{tab:system_comparison}
    \begin{tabular}{cccccc}
        \toprule
        System & Qwen-Plus & Qwen-Plus (w titles) & STORM & CO-STORM & DeepWriter \\
        \midrule
        Model & Qwen-Plus & Qwen-Plus & GPT-4o & GPT-4o & Qwen2-7B \\
        \midrule
        Source & Internal & Internal & search &  search & offline KB \\
        \midrule
        Generation & single turn & single turn &  multi-turn & multi-turn &  multi-turn \\
        \midrule
        web search & no & no & yes & yes & no \\
        \bottomrule
    \end{tabular}
\end{table*}

\paragraph{Benchmark}

We collect a bunch of annual reports spanning across Education, Trade, Health, Refugee and Climate domains, each ranges from several years. Each of them are public accessible, which offers a fair comparison between our proposed offline method and online search methods.
The statistics of these five datasets are given in the Table~\ref{table:evaluation_dataset}. 
As the table shows, the real-world information are diverse and rich of multimodal information, which can offer valuable information.

\begin{table}[!htb]
\begin{tabular}{ccccc}
\toprule
Dataset           & \# PDF & \# pages & \# avg pages & \# images \\
\midrule
Education & $19$    & $8981$    & $472\pm72$      &   $6292$     \\
\midrule
Refugee   & $20$    & $1979$    & $98\pm114$      &   $4644$       \\
\midrule
Climate   & $13$    & $4764$    & $366\pm93$      &   $4232$       \\
\midrule
Health    & $21$    & $2731$    & $130\pm33$      &   $2780$       \\
\midrule
Finance   & $22$    & $5228$    & $237\pm71$      &   $2561$       \\
 \bottomrule
\end{tabular}
\caption{Statistics of evaluation datasets\label{table:evaluation_dataset}}
\end{table}

To run Deepwriter, we manually download the five datasets, serving as the corpus $\mathcal{K}$, then we use MinerU to preprocess them and store the processed results in the database with the corresponding embeddings.
Then, we ask offline methods to write a long article for a specific topic $Q$. 
For online methods, we use the online demo version to generate the corresponding article with the same topic $Q$.

\paragraph{Evaluation}

To evaluate the quality of generated article, following~\cite{shao-etal-2024-assisting, jiang2024unknownunknownsengagedhuman, xi2025omnithinkexpandingknowledgeboundaries}, we adopt 
Prometheus2-7B~\cite{kim2024prometheus} as the judge model to evaluate the quality of the generate article from several dimensions including Interest Level; Coherence and Organization, Relevance and Focus; Coverage. 
The evaluation standard is given in Table~\ref{table:rubric}.

Since Prometheus2-7B can only evaluate the text modality, we opt to use GPT-4o as the judge model to evaluate the coherence of each image and its surrounding paragraphs,

\subsection{Evaluation Results}

\paragraph{Performance}

% TODO: use a radar plot on five datasets, 
\begin{figure}[t]
  \includegraphics[width=\columnwidth]{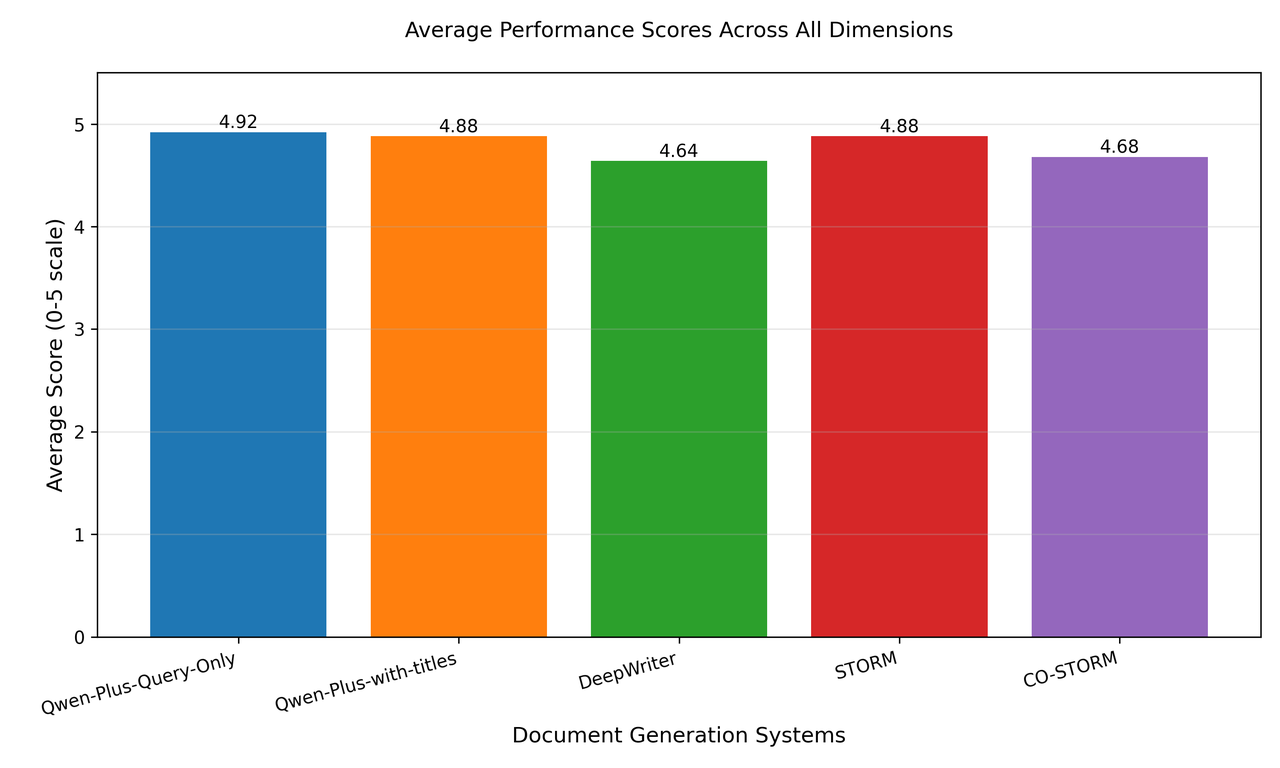}
  \caption{Performance of DeepWriter on WTR dataset against Qwen-Plus, STORM and CO-STORM.}
  \label{fig:DeepWriter_WTR_performance}
\end{figure}

The performance on WTR dataset is given in Fig~\ref{fig:DeepWriter_WTR_performance}.
It can be seen that DeepWriter achieves comparable performance with writing assistants that adopt GPT-4o as foundation models. 
This shows the potential of utilizing compact models to perform long article writing tasks.
We also find that when adding the titles generated by DeepWriter to Qwen-Plus, the coherence improved, which illustrates the effectiveness of task decomposition and planning.

However, there are also limitations when adopting compact models. For example, except for Interest Level and Coherence and Organization, DeepWriter performs worse in the other three dimensions, which reveals the gap between small and large models.

\subsection{Ablation study}

We mainly discuss the impact of different modules to the final performance in this section.

\subsection{Case Study}
In this section, we conduct case studies on success and failure cases to provide deeper insights into DeepWriter's capabilities and limitations.

\section{Conclusion}
In this paper, we introduced DeepWriter, a multimodal writing assistant designed to generate high-quality, long-form documents from a curated, offline knowledge base. We demonstrated that by integrating a hierarchical knowledge structure with a robust pipeline—encompassing task decomposition, multimodal retrieval, and fine-grained citation—DeepWriter can produce factually grounded and coherent articles in specialized domains. Our experiments show that our system achieves competitive performance, particularly in factual accuracy, against strong baselines that rely on larger models or online search, validating the effectiveness of a controlled, offline approach.

\section*{Limitations \& Future Work}
While DeepWriter demonstrates strong performance using offline corpora, several limitations remain that provide opportunities for future research.

\bibliography{aaai2026}

\appendix

\section{Prompt of DeepWriter}
\label{sec:appendix}

\begin{tcolorbox}[colback=blue!5!white,colframe=blue!75!black,title=Query Rewriting Prompt]
Rewrite the following user query in a way that makes it more effective and precise.
The new query should be more specific, focused, and clear, using terminology that is likely to lead to a more accurate understanding of the user's intent. 
Ensure that the rewritten query captures the essence of the user's question while improving its clarity and precision.

Query:
\{query\}

Your rewritten query:
\end{tcolorbox}

\begin{tcolorbox}[colback=blue!5!white,colframe=blue!75!black,title=Task Decomposition Prompt]
You are an expert research assistant. I need to retrieve information from a database to answer the following query:

Query:
\{query\}

Please help me decompose this query into 3-5 more specific, related sub-queries that would help gather comprehensive information to answer the main question. 
These sub-queries should:
\begin{itemize}
    \item Cover different aspects of the main query
    \item Be specific enough for database retrieval
    \item Help gather contextual information needed for a complete answer
    \item Focus on factual information rather than opinions
\end{itemize}

Format your response as a numbered list of sub-queries only. split them with a new line.
\end{tcolorbox}

\begin{tcolorbox}[colback=blue!5!white,colframe=blue!75!black,title=Section Title Generation Prompt]
You are an expert article writer tasked with generating section titles for a comprehensive report. 
Given the following query, generate a list of section titles that would be appropriate for a comprehensive report.

Query:
\{query\}

Instructions:
\begin{enumerate}
    \item The section titles should follow the human-like structure of a report.
    \item The content of the section should be related to the query.
    \item The section titles should from general to specific. Like: Background, Analysis, Viewpoints.
    \item  split the section titles by new line such that each line contains exactly one section title. Example:
    \begin{itemize}
        \item Background
        \item Analysis
        \item Viewpoints
    \end{itemize}
\end{enumerate}

Your section titles:
\end{tcolorbox}

\begin{tcolorbox}[colback=blue!5!white,colframe=blue!75!black,title=Section Draft Prompt]
You are an expert research writer tasked with creating a section draft for a section of a comprehensive report.

Query:
\{query\}

Section Title:
\{section\_title\}

Relevant Documents:
\{relevant\_docs\}

Instructions:
\begin{enumerate}
    \item  Analyze the query and section title and figure out what should be included in this section.
    \item Create a rough draft for writing this section that covers the information revealed by relevant documents.
    \item Be simple and concise.
    \item DO NOT add references to the draft.
    \item Try to avoid using bullets and subsections, just synthesize the information in a natural way.
\end{enumerate}

Your draft should provide a high-level perspective on how to approach writing this section effectively.
Focus on organization and content strategy rather than specific wording.

Provide your draft below:
\end{tcolorbox}

\begin{tcolorbox}[colback=blue!5!white,colframe=blue!75!black,title=Document Clustering Prompt]
You are an expert document classifier. Your task is to classify the given document into the most appropriate section based on its content and relevance to the query.

Query:
\{query\}

Document:
\{doc\}

Available sections:
\{sections\}

Instructions:
\begin{enumerate}
    \item Carefully analyze the document content in relation to the query
    \item Consider how the information would fit into a structured report addressing the query
    \item Choose EXACTLY ONE section from the available sections where this document would be most appropriate
    \item Return ONLY the name of the chosen section, with no additional text or explanation
\end{enumerate}

Your classification (return only the section name):
\end{tcolorbox}

\begin{tcolorbox}[colback=blue!5!white,colframe=blue!75!black,title=Section Content Generation Prompt]
You are an expert research writer tasked with generating high-quality content for a specific section of a comprehensive report.

Query:
\{query\}

Section Title:
\{section\_title\}

Section Draft:
\{section\_draft\}

Relevant Documents:
\{relevant\_docs\}

Content Already Written in Previous Sections:
\{already\_written\}

Instructions:
\begin{enumerate}
    \item Generate detailed, well-structured content for the "{section\_title}" section that directly addresses the query
    \item Incorporate information from the relevant documents, synthesizing and analyzing the data
    \item Ensure continuity with content already written in previous sections
    \item Use an academic, professional tone appropriate for a research report
    \item Be thorough but concise, focusing on information that is most relevant to the query
    \item Avoid repetition of content already covered in previous sections
    \item Do not include title in any level just write the content
\end{enumerate}

Your content should:
\begin{itemize}
    \item  Present factual information directly derived from the relevant documents
    \item Synthesize and organize information from multiple sources
    \item Maintain neutrality when presenting evidence and data
\end{itemize}
\end{tcolorbox}

\begin{tcolorbox}[colback=blue!5!white,colframe=blue!75!black,title=Summarization Prompt]
You are an expert summarizer. Your task is to create a concise and accurate summary of the following content in relation to a specific query.

The summary should:
\begin{enumerate}
    \item Capture the main points and key information relevant to the query
    \item Highlight the relationship between the content and the query, if there is no relationship, return "None"
    \item Maintain the original meaning and intent
    \item Be clear and coherent
    \item Be no more than 30 percent of the original length
\end{enumerate}

Query:
\{query\}

Content to summarize:
\{doc\}

Provide your summary below, focusing on aspects that address the query:
\end{tcolorbox}

\section{Key Subroutines}
Algorithm~\ref{alg:task_decomposition} shows the algorithm on task decomposition. Algorithm~\ref{alg:placement_optimization} demonstrates the multimodal placement optimization. 
Algorithm~\ref{alg:citation_generation} reveals how fine-grained citation generation works.

\begin{algorithm}[H]
\caption{Task Decomposition\label{alg:task_decomposition}}
\begin{algorithmic}[1]
\REQUIRE Query $Q'$
\ENSURE Subtasks $\mathcal{T} = \{\mathcal{T}_{\text{fact}}, \mathcal{T}_{\text{data}}, \mathcal{T}_{\text{point}}\}$

\STATE $\mathcal{T}_{\text{fact}} \leftarrow \emptyset$
\STATE $\mathcal{T}_{\text{data}} \leftarrow \emptyset$
\STATE $\mathcal{T}_{\text{point}} \leftarrow \emptyset$

\STATE \textbf{Extract Facts}
\FOR{each entity $e$ in $Q'$}
    \IF{$\text{IsAmbiguous}(e)$}
        \STATE $\mathcal{T}_{\text{fact}} \leftarrow \mathcal{T}_{\text{fact}} \cup \{\text{Clarify}(e)\}$
    \ENDIF
\ENDFOR

\STATE \textbf{Extract Data Requirements}
\FOR{each quantitative term $q$ in $Q'$}
    \STATE $\mathcal{T}_{\text{data}} \leftarrow \mathcal{T}_{\text{data}} \cup \{\text{Quantify}(q)\}$
\ENDFOR

\STATE \textbf{Extract Points}
\STATE $\mathcal{T}_{\text{point}} \leftarrow \text{IdentifyKeyArguments}(Q')$

\RETURN $\mathcal{T}$

\end{algorithmic}
\end{algorithm}

\begin{algorithm}[H]
\caption{Multimodal Placement Optimization\label{alg:placement_optimization}}
\begin{algorithmic}[1]
\REQUIRE Visual elements $\mathcal{M}$, Document $P$
\ENSURE Optimized placement positions

\STATE $\text{Sim} \leftarrow \emptyset$
\FOR{each visual $m_k \in \mathcal{M}$}
    \FOR{each paragraph $p_i \in P$}
        \STATE $\text{score}_{k,i} \leftarrow \text{GME\_Similarity}(m_k, p_i)$
        \STATE $\text{Sim}[k][i] \leftarrow \text{score}_{k,i}$
    \ENDFOR
\ENDFOR

\STATE \textbf{Placement Optimization}
\FOR{each visual $m_k \in \mathcal{M}$}
    \STATE $\text{best\_pos} \leftarrow \arg\max_i \text{Sim}[k][i]$
    \STATE $\text{constraints} \leftarrow \text{CheckFlowConstraints}(m_k, \text{best\_pos}, P)$
    \IF{$\text{constraints.satisfied}$}
        \STATE $\text{pos}_k \leftarrow \text{best\_pos}$
    \ELSE
        \STATE $\text{pos}_k \leftarrow \text{FindAlternativePosition}(m_k, \text{constraints})$
    \ENDIF
\ENDFOR

\RETURN $\{\text{pos}_k\}_{k=1}^{|\mathcal{M}|}$

\end{algorithmic}
\end{algorithm}

\begin{algorithm}[H]
\caption{Fine-grained Citation Generation\label{alg:citation_generation}}
\begin{algorithmic}[1]
\REQUIRE Claim $c$, Retrieved content $\mathcal{R}$
\ENSURE Citation with precise source location

\STATE $\text{best\_match} \leftarrow \emptyset$
\STATE $\text{highest\_score} \leftarrow 0$

\FOR{each retrieved item $r \in \mathcal{R}$}
    \STATE $\text{score} \leftarrow \text{SemanticSimilarity}(c, r.\text{content})$
    \IF{$\text{score} > \text{highest\_score}$}
        \STATE $\text{highest\_score} \leftarrow \text{score}$
        \STATE $\text{best\_match} \leftarrow r$
    \ENDIF
\ENDFOR

\STATE $\text{citation} \leftarrow \text{CreateCitation}(\text{best\_match})$
\STATE $\text{citation.level} \leftarrow \text{DetermineGranularity}(\text{best\_match})$

\IF{$\text{citation.level} = \text{'document'}$}
    \STATE $\text{citation.reference} \leftarrow \text{best\_match.filename}$
\ELSIF{$\text{citation.level} = \text{'paragraph'}$}
    \STATE $\text{citation.reference} \leftarrow \text{best\_match.filename} + \text{':'} + \text{best\_match.paragraph\_id}$
\ELSE
    \STATE $\text{citation.reference} \leftarrow \text{best\_match.filename} + \text{':'} + \text{best\_match.sentence\_id}$
\ENDIF

\RETURN $\text{citation}$

\end{algorithmic}
\end{algorithm}

\section{LLM Evaluation}
See Table~\ref{table:rubric} for details.

\begin{table*}
\centering
\begin{tabular}{ll} 
\toprule
Criteria Description & \textbf{Interest Level}: How engaging and thought-provoking is the article?                                                                                                               \\
Score 1 Description  & Not engaging at all; no attempt to capture the reader's attention.                                                                                                                        \\
Score 2 Description  & Fairly engaging with a basic narrative but lacking depth.                                                                                                                                 \\
Score 3 Description  & Moderately engaging with several interesting points.                                                                                                 \\
Score 4 Description  & Quite engaging with a well-structured narrative and noteworthy points that frequently capture and retain attention.                                                                       \\
Score 5 Description  & Exceptionally engaging throughout, with a compelling narrative that consistently stimulates interest.                                                                                     \\ 
\midrule
Criteria Description & \textbf{Coherence and Organization}: Is the article well-organized and logically structured?                                                                                              \\
Score 1 Description  & Disorganized; lacks logical structure and coherence.                                                                                                                                      \\
Score 2 Description  & Fairly organized; a basic structure is present but not consistently followed.                                                                                                             \\
Score 3 Description  & Organized; a clear structure is mostly followed with some lapses in coherence.                                                                                                 \\
Score 4 Description  & Good organization; a clear structure with minor lapses in coherence.                                                                                                                      \\
Score 5 Description  & Excellently organized; the article is logically structured with seamless transitions and a clear argument.                                                                                \\ 
\midrule
Criteria Description & \textbf{Relevance and Focus}: Does the article stay on topic and maintain a clear focus?                                                                                                  \\
Score 1 Description  & Off-topic; the content does not align with the headline or core subject.                                                                                                                  \\
Score 2 Description  & Somewhat on topic but with several digressions; the core subject is evident but not consistently adhered to.                                                                              \\
Score 3 Description  & Generally on topic, despite a few unrelated details.                                                                                                                                      \\
Score 4 Description  & Mostly on topic and focused; the narrative has a consistent relevance to the core subject with infrequent digressions.                                                                    \\
Score 5 Description  &Exceptionally focused and entirely on topic; the article is tightly centered on the subject, with every piece of information contributing to a comprehensive understanding of the topic.  \\ 
\midrule
Criteria Description & \textbf{Broad Coverage}: Does the article provide an in-depth exploration of the topic and have good coverage?                                                                            \\
Score 1 Description  & Severely lacking; offers little to no coverage of the topic's primary aspects, resulting in a very narrow perspective.                                                                    \\
Score 2 Description  & Partial coverage; includes some of the topic's main aspects but misses others, resulting in an incomplete portrayal.                                                                      \\
Score 3 Description  & Acceptable breadth; covers most main aspects, though it may stray into minor unnecessary details or overlook some relevant points.                                                        \\
Score 4 Description  & Good coverage; achieves broad coverage of the topic, hitting on all major points with minimal extraneous information.                                                                     \\
Score 5 Description  & Exemplary in breadth; delivers outstanding coverage, thoroughly detailing all crucial aspects of the topic without including irrelevant information.                                     \\
\bottomrule
\end{tabular}

\caption{Scoring rubrics on a 1-5 scale for the evaluator LLM.}
\label{table:rubric}
\end{table*}

\end{document}